# Data Augmentation using Transformers and Similarity Measures for Improving Arabic Text Classification

Dania Refai
*Department of Computer Science*
*Princess Sumaya University for Technology (PSUT)*
Amman, Jordan
dania.refai@hotmail.com

Saleh Abu-Soud
*Department of Data Science*
*Princess Sumaya University for Technology (PSUT)*
Amman, Jordan
abu-soud@psut.edu.jo

Mohammad Abdel-Rahman
*Department of Data Science*
*Princess Sumaya University for Technology (PSUT)*
Amman, Jordan
mo7ammad@vt.edu

**ABSTRACT** The performance of learning models heavily relies on the availability and adequacy of training data. To address the dataset adequacy issue, researchers have extensively explored data augmen- tation (DA) as a promising approach. DA generates new data instances through transformations applied to the available data, thereby increasing dataset size and variability. This approach has enhanced model performance and accuracy, particularly in addressing class imbalance problems in classification tasks. However, few studies have explored DA for the Arabic language, relying on traditional approaches such as paraphrasing or noising-based techniques. In this paper, we propose a new Arabic DA method that employs the recent powerful modeling technique, namely the AraGPT-2, for the augmentation process. The generated sentences are evaluated in terms of context, semantics, diversity, and novelty using the Euclidean, cosine, Jaccard, and BLEU distances. Finally, the AraBERT transformer is used on sentiment classification tasks to evaluate the classification performance of the augmented Arabic dataset. The experiments were conducted on four sentiment Arabic datasets: AraSarcasm, ASTD, ATT, and MOVIE. The selected datasets vary in size, label number, and unbalanced classes. The results show that the proposed methodology enhanced the Arabic sentiment text classification on all datasets with an increase in F1 score by 4% in AraSarcasm, 6% in ASTD, 9% in ATT, and 13% in MOVIE.

## I. INTRODUCTION

Natural language processing (NLP) is a branch of artificial intelligence that aims to teach computers to process and analyze large volumes of natural language data [1], [2]. Machine learning and deep learning have made significant advances in recent years, particularly in the NLP field [3]. However, the learning model in machine learning systems is highly dependent on data, making it difficult to obtain a large amount of labeled data, particularly in domains such as education and healthcare [4].

Data augmentation (DA) has emerged as a promising approach to address the issue of dataset adequacy [5]–[7]. DA increases the number of training data instances by performing various transformations on actual data instances to generate new and representative data instances, thereby improving the model's efficiency and prediction accuracy [6]. Additionally, DA helps to minimize overfitting and solve the class imbalance issue in classification learning techniques [8]. Although DA is well-established in computer vision and speech recognition, it is not a common practice in the NLP field [8]. Traditional methods of increasing text data are costly and time-consuming, particularly when there are not enough resources to support the augmentation process, such as language dictionaries or databases of synonyms for the chosen dataset. Furthermore, not all augmentation methods are applicable to all languages, as certain transformations may make the sentence grammatically or semantically incorrect [7], [9].

Using pre-trained transformer models in DA can help to overcome these limitations [10]. Transformer models have proven effective in various NLP tasks, including text sum- marization, translation, generation, and question-answering systems [11]. Additionally, employing transformer models in the DA process preserves the text context and dependencies between the sentence words, thus solving the issues associ- ated with traditional augmentation methods [12]–[15].

However, it is essential to assess the quality of augmented text from various perspectives, including context, semantics, diversity, and novelty [16]. Text-similarity metrics such as Euclidean [17], cosine [18], Jaccard [19], and BLEU [20] measures can be used to evaluate the quality of augmented text [16].

After conducting a thorough review of existing literature, it is evident that various DA techniques have been implemented in different languages, with a focus on English. These techniques have proven to be effective in enhancing English language learning, and fall into two categories: paraphrasing techniques using thesauruses [6], translation [21], or transformers [13], and techniques that add noise to sentence words, such as swapping [22], deletion [4], insertion [23], and substitution [24].

Despite Arabic being the fifth most spoken language globally [25] and experiencing significant growth of digital Arabic content on the Internet [26], there is a significant gap in research when it comes to DA for Arabic data. One of the challenges in Arabic DA is the language's unique characteristics [27], which make it difficult to accurately augment textual data using conventional methods such as paraphrasing or noising-based techniques [28]–[31]. Additionally, the current body of research lacks studies that employ DA techniques on Arabic data and utilize all similarity measures to evaluate the quality of the generated sentence, which is crucial for effective language learning. Therefore, there is a pressing need for further exploration of the potential of using Arabic transformers, such as AraGPT2 [32] and AraBERT's [33], in DA for Arabic data, along with a comprehensive assessment of generated sentence quality [14], [15]. Combining transformers and similarity measures could solve the challenges of Arabic DA and improve the accuracy of generated sentences, which can, in turn, enhance model learning outcomes.

*MAIN CONTRIBUTIONS:*

Our main contributions in this paper can be summarized as follows:

- A novel approach for Arabic textual data augmentation. Our method harnesses the capabilities of recent powerful tools based on the transformer's architecture. Specifically, our method utilizes AraGPT-2's text generation task [32] for paraphrasing in the augmentation process.
- Different text evaluation metrics are used to evaluate the generated sentences from our approach in terms of context, semantics, diversity, and novelty. Specifically, the Euclidean, cosine, Jaccard, and BLEU distances are used.
- Sentiment classification is performed on the augmented Arabic dataset using the AraBERT [33] transformer, and the effects of DA on classification performance have been examined.

*PAPER ORGANIZATION:*

The rest of the paper is organized as follows. In Section II, we provide a literature review. Our proposed methodology is explained in Section III. In Section IV, we introduce the sentiment datasets used for evaluating our proposed methodology. Section V discusses the experiments conducted to assess the robustness of the proposed approach along with their results. Comparisons with related works are also provided in Section V. Finally, our conclusions are summarized in Section VI.

## II. LITERATURE REVIEW

Language transformer models, such as GPT-3 [32], belong to a class of neural network architectures that have revolutionized NLP in recent years. These models are typically pre-trained on large corpora of text data to learn general language patterns and relationships between words. One of the most influential architectures for language transformers is the transformer model, introduced in [10]. Transformers have been predominantly trained on English text, which has led to their success in various NLP tasks [12]–[15]. However, researchers have recently started adapting these powerful models for other languages, including Arabic [34]–[38]. To do so, they pre-train the transformer models on large Arabic textual datasets, such as AraBERT [33], AraGPT-2 [32], and AraElectra [39]. Pre-training on Arabic text allows these models to learn language patterns and relationships specific to the Arabic language, which makes them highly effective for Arabic NLP tasks. Despite their effectiveness in preserving context in natural language [10], studies on Arabic have not yet explored using language transformer models as an augmentation technique.

Furthermore, to ensure that augmented data improves performance without altering the meaning of the original data, it is crucial to evaluate its quality before incorporating it into the augmented Arabic dataset [9]. Evaluating sentences in context and assessing their quality in terms of semantics, diversity, novelty, and other factors is necessary to effectively evaluate the augmented data [16]. While some researchers have used the Jaccard similarity metric [19] to evaluate the novelty and diversity of generated sentences in Arabic DA processes before adding them to the dataset [40], a more comprehensive evaluation is required that considers various aspects of sentence quality, such as context, semantics, diversity, and novelty [16].

Text classification is a widely researched area in NLP, with much attention given to languages like English and Spanish [41]. However, Arabic language text classification has received a different level of attention, mainly due to the unique characteristics of the language that require different methodologies [42], [43]. While existing classification methods for Arabic text are still limited, transformers have emerged as a promising tool for improving Arabic DA tasks, including text classification [43]. Furthermore, various Arabic studies have employed advanced models, such as AraBERT, MARRBERT, ArBERT, QARiB, AraBERTv02, GigaBERT, ArabicBERT, and mBERT to evaluate their augmented Arabic datasets using classification tasks [34], [35], [44]. For instance, authors in [45] used MARBERT and QARiB to distinguish between human-generated and fake-generated tweets with high accuracy [36], [46].

The proposed DA techniques and methods can be broadly classified into two main categories: paraphrasing-based and noising-based [4], [9], [47], [48]. Regarding paraphrasing-based techniques, recent studies have employed transformer models as an augmentation process, demonstrating their efficiency in various NLP tasks, including text summarization, translation, classification, generation, named entity recognition, and question-answering systems [11]. Although using transformer models in augmentation preserves the text context, it is essential to note that the augmented text should be evaluated from various aspects, including context, semantics, diversity, and novelty [16]. Text-similarity measurements can be used to check the quality of the augmented sentences [40]. Various text similarity metrics can be used, including Euclidean distance [17], cosine distance [18], Jaccard distance [19], and BLEU distance [20].

While few studies have focused on augmenting Arabic data [28]–[31], [49], some have used the current DA noising and paraphrasing-based approaches without employing the transformer's powerful models as augmentation techniques [29]–[31], [50]. Other studies have employed transformers to evaluate the augmented Arabic dataset [34], [35]. A few studies have considered classification tasks using the AraBERT transformer and achieved the best results in Arabic text classification [34]–[36]. Recently, one study considered the Jaccard metric to evaluate the novelty of the generated sentences in Arabic text [40]. As mentioned earlier, the generated sentence should be evaluated from different aspects, such as context, semantics, diversity, and novelty [16]. Table 1 summarizes the DA techniques used in the Arabic language and their results.

DA techniques that leverage transformers and similarity metrics have shown significant advantages in English textual data classification [12]–[15]. However, a noticeable gap exists in current research regarding utilizing transformers and similarity metrics for Arabic textual data augmentation and classification processes. As depicted in Fig. 1, previous Arabic studies mainly focused on using transformers exclusively for classification, with only one study employing Jaccard similarity to assess the generated sentences. Our research proposes a novel approach encompassing three key aspects to address this gap. Firstly, we introduce a groundbreaking methodology that harnesses the power of recent transformer-based tools for data augmentation in Arabic. Secondly, we adopt diverse text evaluation metrics, including Euclidean, cosine, Jaccard, and BLEU distances, to thoroughly assess the generated sentences, focusing on context, semantics, diversity, and novelty. Additionally, our research includes sentiment classification on the augmented Arabic dataset, enabling us to explore the impact of data augmentation on classification performance. By encompassing these key elements, our methodology effectively bridges the research gap and significantly advances the field of DA techniques in Arabic, as shown in the dashed box in Fig. 1.

## III. METHODOLOGY
In this section, we propose a three-phase empirical approach for Arabic DA. In the first phase, we use the AraGPT-2-

**TABLE 1.** A summary of the Arabic DA techniques and their final findings.

| DA Technique Study | Dataset Name and Macro F1 Results | | | |
| | Ara-Sarcasm | Twitter data [51] | Product reviews [52] | Arab Gloss-BERT dataset |
| --- | --- | --- | --- | --- |
| Noising-based DA techniques, including word replacement, insertion, and mix between them [53] | 0.46 | — | — | — |
| Noising-based DA techniques including word replacement, insertion, and mix between them [31] | 0.75 | — | — | — |
| Noising-based DA techniques including merging an external dataset with AraSarcesm dataset [39] | 0.52 | — | — | — |
| Noising-based DA techniques including manually expanding the dataset [36] | — | 96.0 | — | — |
| Paraphrasing-based DA techniques using language rules [29] | — | — | 0.65 | — |
| Paraphrasing-based DA technique using Arabic-English Arabic back-translation [50] | — | — | — | between 65.0 and 89.0 |

base [32] pre-trained model to generate Arabic text from the given dataset records. This results in a new dataset that contains the generated Arabic text from the transformer (i.e., the AraGPT-2-base). In the second phase, we add new records to the given dataset by employing the similarity measures, namely, the Euclidean [17], cosine [18], Jaccard [19], and BLEU [20] distances. The augmentation process depends on (i) the similarity thresholds and (ii) the selected class labels for the data to be augmented. The third phase comes as a complementary phase, which assists in evaluating the performance of the text classification process on the newly created dataset (i.e., the augmented dataset). Fig. 2 illustrates the general phases of the adopted methodology. In the following subsections, we explain the three phases of the DA process.

### A. PHASE 1: ARABIC TEXT DATA GENERATION USING TRANSFORMERS
In this phase, the dataset to be augmented is first loaded. Then, a transformer that can generate Arabic text is created (AraGPT-2-base [32] is used in this paper) along with initializing the similarity functions needed to calculate the similarity between the old Arabic text supplied to the transformer and the newly generated text from the transformer (i.e., to/from the AraGPT-2-base transformer). Subsequently,

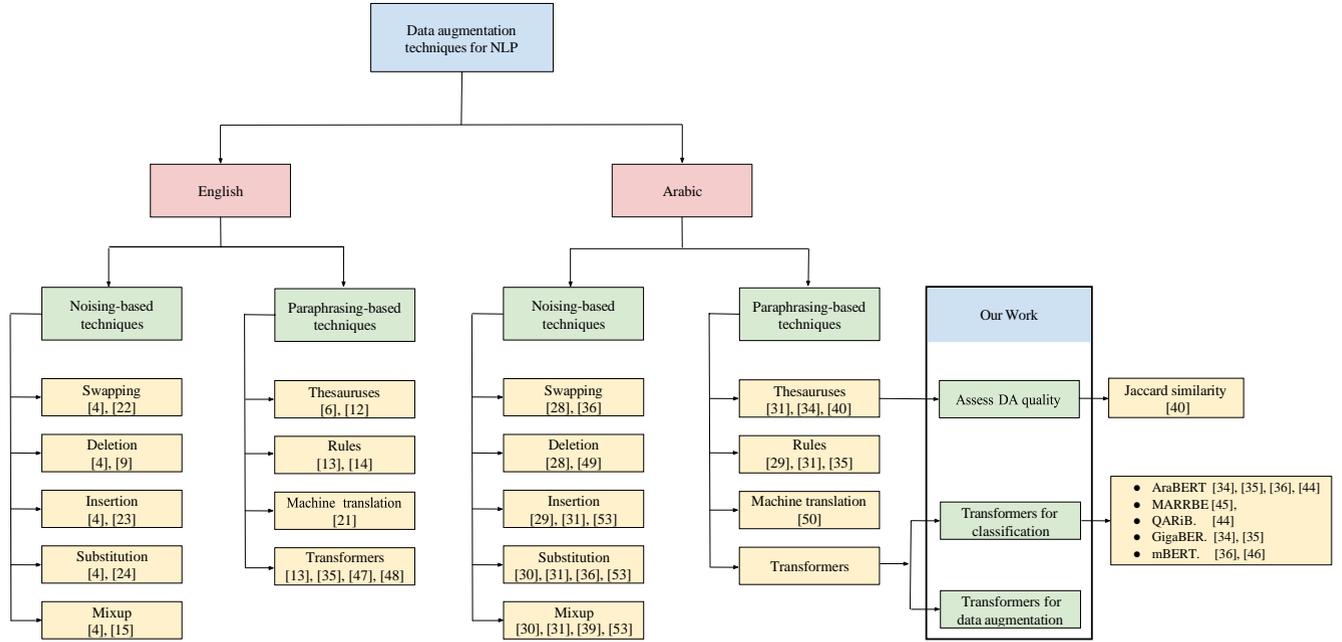

**FIGURE 1.** Taxonomy of DA techniques for NLP.

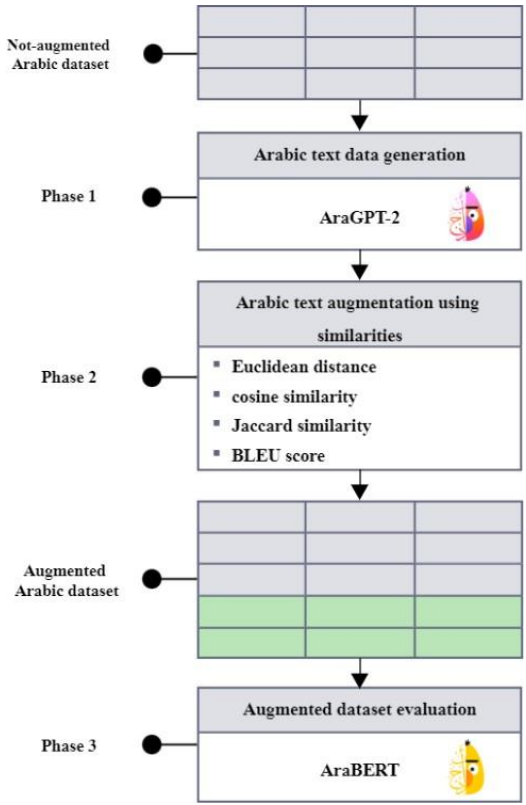

**FIGURE 2.** The main phases of the adopted methodology.

for each record in the given dataset's records:

- First, the given Arabic text in the record is preprocessed using the provided preprocessor of the selected transformer, which is the AraBERT preprocessor [30], [33], [35].

- Second, a need to calculate the word embedding that represents the given Arabic text would take place. Such a sub-step is needed since the similarity functions deal with numerical representations (i.e., vectors) rather than the abstract Arabic text representation to calculate the distances between the objects for comparing them. Hence, in this paper, we used BERT word embedding for computing the word embedding [54].

- Third, the similarity between the numerical representation (i.e., the words embedding) of the given Arabic text in the record and the newly generated one is calculated with the selected similarity functions (Euclidean, cosine, Jaccard, and BLEU distances).

- Finally, all the computed and generated information were collected within the current loop (the given Arabic text, the related class label, the newly generated text, all text of the given Arabic text combined with the generated one, the embedding representation, and the similarities' values), and is appended to the current record. Moreover, such a record is added to the final dataset to be exported upon finishing this phase, along with the original class label related to the current record being processed.

Phase 1 of the proposed solution is summarized in Fig. 3, which provides a visual overview of the steps involved. The corresponding algorithm is presented in Algorithm 1, which outlines the sequence and flow of operations. To further clarify the workflow, we include an illustrative example in Fig. 4, based on a single record of Arabic text. Together, these resources offer a clear and comprehensive description of Phase 1 of our approach.

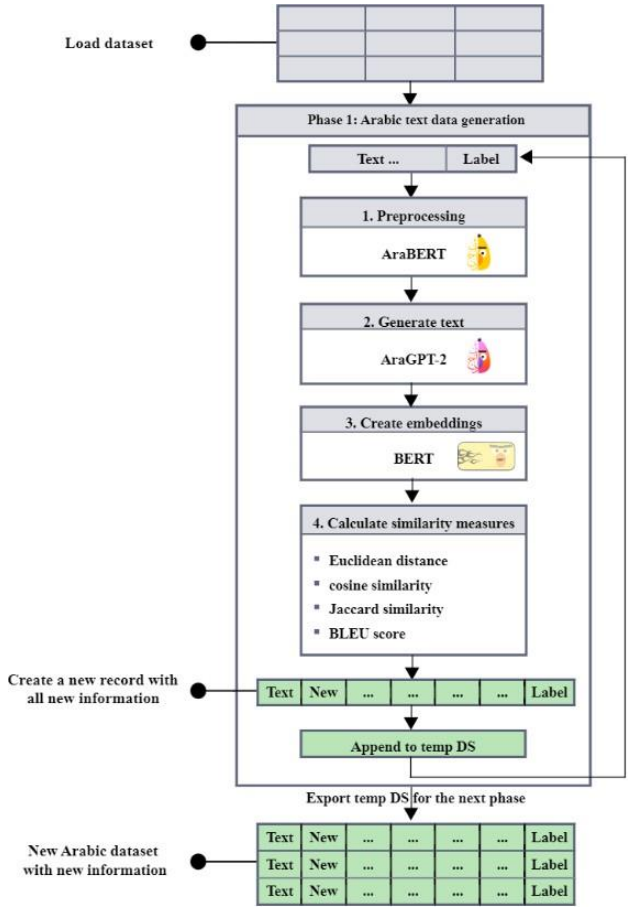

**FIGURE 3.** Methodology steps contained within phase 1.

| Phase No. | Steps | Sentence Modifications |
|---|---|---|
| Phase 1 | Original Sentence | طموحي تضمن أن أكمل تعليمي وأحصل على الشهادات العليا بدرجة 100%، لأسعد أمي وأبي 😊 ومن ثم نفسي |
| | Preprocessed Sentence | طموحي تضمن أن أكمل تعليمي وأحصل على الشهادات العليا بدرجة لأسعد أمي وأبي 😊 ومن ثم نفسي |
| | Generated Sentence | وعائلتي وأصدقائي وأحبابي في كل مكان |
| | All text | طموحي تضمن أن أكمل تعليمي وأحصل على الشهادات العليا بدرجة لأسعد أمي وأبي 😊 ومن ثم نفسي وعائلتي وأصدقائي وأحبابي في كل مكان |

**FIGURE 4.** Augmentation illustrative example.

### B. PHASE 2: ARABIC DATASET AUGMENTATION USING SIMILARITIES

In this phase, the generated dataset from Phase 1 is processed to generate one final dataset that contains the new augmented records. Generating this final dataset requires two significant decisions: (i) selecting the classes to be augmented and (ii) selecting a threshold (i.e., similarity-desired value) to decide

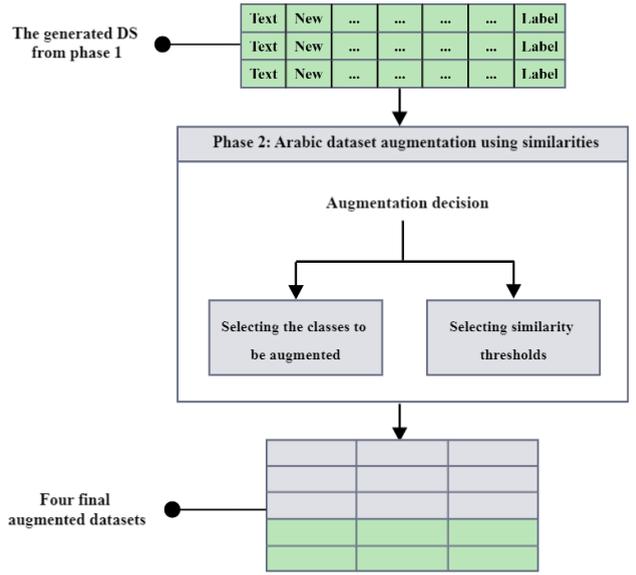

**FIGURE 5.** Methodology steps contained within phase 2.

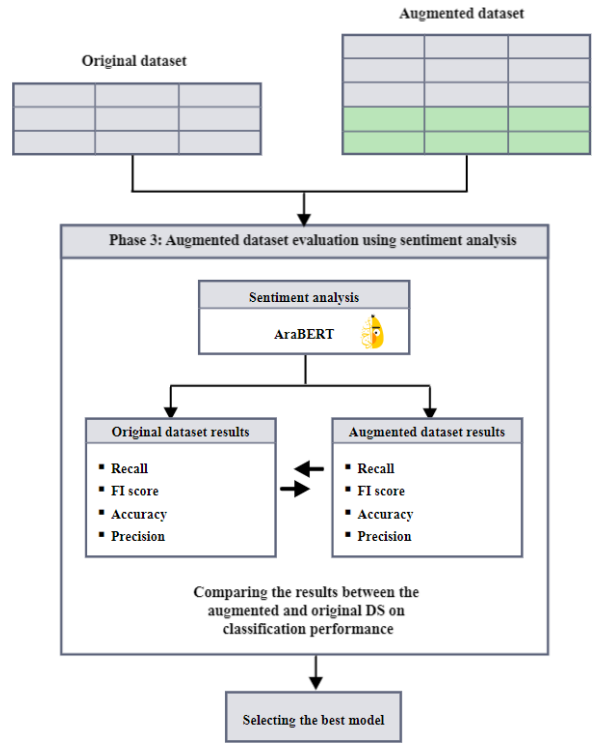

**FIGURE 6.** Methodology steps contained within phase 3.

---

**Algorithm 1** Arabic Text Data Generation.

---

**Result :** Generated dataset with similarity measures.
**Input :** Original dataset [original sentence, label]
**Output:** Temp dataset [original sentence, generated sentence, all text, original sentence embeddings, generated sentence embeddings, Euclidean similarity, cosine similarity, Jaccard similarity, BLEU similarity].

**for** *each record in original dataset* **do**

    1) Preprocessing (original sentence).

    2) Generated sentence ← generates text (original sentence).

    3) Original sentence embeddings ← create embeddings (original sentence).

    4) Generated sentence embeddings ← create embeddings (generated sentence).

    5) Euclidean similarity ← calculate Euclidean (original sentence, generated sentence).

    6) cosine similarity ← calculate cosine (original sentence, generated sentence).

    7) Jaccard similarity ← calculate Jaccard (original sentence, generated sentence).

    8) BLEU similarity ← calculate BLEU (original sentence, generated sentence).

    9) All text ← combine text (original sentence, generated sentence).

    10) Temp dataset.add (original sentence, generated sentence, all text, original sentence embeddings, generated sentence embeddings, and similarities).

**end**
Export temp dataset.

---

**Algorithm 2** Arabic Dataset Augmentation using Similarities.

---

**Result :** Augmented datasets based on similarity measures.
**Input :** Temp dataset.
**Output:** Euclidean augmented dataset, cosine augmented dataset, Jaccard augmented dataset, BLEU augmented dataset.

**Step 1:** Create empty datasets.

    1) Euclidean augmented dataset.

    2) cosine augmented dataset.

    3) Jaccard augmented dataset.

    4) BLEU augmented dataset.

    5) Augmented classes.

**Step 2:** Calculate similarity thresholds for each measure.

**for** *all records in temp dataset* **do**

    1) Euclidean threshold ← average (Euclidean similarity).

    2) cosine threshold ← average (cosine similarity).

    3) Jaccard threshold ← average (Jaccard similarity).

    4) BLEU threshold ← average (BLEU similarity).

**end**

**Step 3:** Augment datasets.

**for** *each record in temp dataset* **do**

    **if** *Euclidean similarity $\geq$ Euclidean threshold* **then**

        Euclidean augmented dataset.add (all text).

        Augmented classes.add (label).

    **end**

    **if** *cosine similarity $\geq$ cosine threshold* **then**

        cosine augmented dataset.add (all text).

        Augmented classes.add (label).

    **end**

    **if** *Jaccard similarity $\geq$ Jaccard threshold* **then**

        Jaccard augmented dataset.add (all text).

        Augmented classes.add (label).

    **end**

    **if** *BLEU similarity$\geq$BLEU threshold* **then**

        BLEU augmented dataset.add (all text).

        Augmented classes.add (label).

    **end**

**end**

**Step 4:** Export datasets:

    1) Euclidean augmented dataset.

    2) cosine augmented dataset.

    3) Jaccard augmented dataset.

    4) BLEU augmented dataset.

---

the selection process of the newly generated text as a new record in the new dataset along with the related class label. Accordingly, the similarity threshold percentage for each similarity metric is calculated by taking the average for each similarity column value from the (exported temp DS) from Phase 1. Fig. 5 summarizes the implemented steps to achieve these two significant decisions. Furthermore, the sequence and flow for Phase 2 operations are depicted in Algorithm 2. Consequently, the final collected dataset upon this selection strategy is exported for the next phase (i.e., Phase 3).

### C. PHASE 3: AUGMENTED DATASET EVALUATION USING SENTIMENT ANALYSIS

In this phase, the final augmented datasets are evaluated using sentiment analysis [55] since all the selected Arabic datasets are classified with the sentiment of the text [49], [51]–[53]. To conclude this evaluation, the model of the original dataset (i.e., the dataset before augmentation) with a selected classifier is needed to find the final classification performance results (for instance, the recall, F1, accuracy, etc.). Then, compare the obtained results with the results found in the same classification process. In this context, Fig. 6 depicts the combined classification process steps.

In this view, the AraBERT base Twitter classifier named "aubmindlab/bert-base-arabertv02-twitter" [32] is used. However, the data sets' splits for the classification processes were selected to be 80% for training and 20% for testing on both types of the datasets at hand (i.e., the original datasets before augmentation and the augmented datasets). Meanwhile, the $K$-fold cross-validation approach [56] was adopted for validating and finding the best model's hyper-parameters for the classification process of the sentiment contained within the given data sets' types.

### IV. DATESET SELECTED

The Arabic language is highly morphologically rich, with one Arabic word having multiple meanings and shapes, which requires a comprehensive understanding of the language [27]. To ensure that our proposed approach is robust and applicable to a range of scenarios, we have considered multiple datasets that cover various aspects of the Arabic language, as described in Table 2. Our selection includes diverse Arabic dialects and cases with random examples that may affect the proposed approach in this paper. This ensures the correctness of the proposed approach and avoids limitations to a single dataset with limited examples and fewer characteristics of the Arabic language [34], [40], [42].

**TABLE 2.** Description of the data sets considered for experimentation.

| Dataset Short Name | Description |
|---|---|
| AraSarcasm-v1 [49] | AraSarcasm is a new dataset for detecting sarcasm in Arabic. The dataset was built by adding sarcasm and dialect labels to previously accessible Arabic sentiment analysis datasets (SemEval 2017 and ASTD). There are 10, 547 tweets in the dataset, with 1, 682 (16%) of them being snarky [49] |
| ASTD [57] | ASTD is a dataset that is collected from tweets after being filtered and annotated by the authors to be an Arabic social sentiment analysis dataset gathered from Twitter. The final number of records contained in this dataset is 3224 records. The dataset's number of classes was initially four classes (NEG, POS, NEUTRAL, OBJ). Nevertheless, the authors consider the dataset with the records labeled (NEG, POS, NEUTRAL) for the experimentation [57] |
| ATT [51] | Another Arabic dataset for the reviews expresses the attraction sentiment of the travelers. Moreover, such a dataset was collected from TripAdvisor.com, and it has 2154 records labeled with positive and negative classes [51] |
| MOVIE [51] | Another dataset is scrapped from TripAdvisor.com and contains 1524 records. It contained three classes, namely, positive, negative, and neutral classes [51] |

**TABLE 3.** Data sets considered for experimentation in the proposed solution.

| Dataset Name | Record No. | Class Labels' Information | | |
|---|---|---|---|---|
| | | Label Name | No. Instances | Ratio (%) |
| Ara-Sarcasm | 10545 | POSITIVE | 1678 | 15.91% |
| | | NEUTRAL | 5339 | 50.63% |
| | | NEGATIVE | 3528 | 33.46% |
| ASTD | 3221 | POSITIVE | 776 | 24.09% |
| | | NEUTRAL | 805 | 24.99% |
| | | NEGATIVE | 1640 | 50.92% |
| ATT | 2151 | POSITIVE | 81 | 3.77% |
| | | NEGATIVE | 2070 | 96.23% |
| MOVIE | 1517 | POSITIVE | 966 | 63.68% |
| | | NEUTRAL | 170 | 11.21% |
| | | NEGATIVE | 381 | 25.12% |

**TABLE 4.** Class labels to be augmented and similarity thresholds.

| Dataset Name | Class Labels | Similarities | | | |
|---|---|---|---|---|---|
| | | Euclidean | cosine | Jaccard | BLEU |
| Ara-Sarcasm | NEGATIVE, POSITIVE | 0.327 | 0.835 | 0.265 | 0.316 |
| ASTD | NEUTRAL, POSITIVE | 0.331 | 0.852 | 0.362 | 0.394 |
| ATT | NEGATIVE | 0.193 | 0.865 | 0.208 | 0.447 |
| MOVIE | NEGATIVE, NEUTRAL | 0.028 | 0.904 | 0.0003 | 0.071 |

## V. EXPERIMENTAL RESULTS AND DISCUSSION

### A. EXPERIMENT DATA

As mentioned earlier, we have selected several sentiment Arabic datasets with different characteristics, including dataset size, label number, and unbalanced classes, to evaluate the impact of the proposed augmentation methodology on classification performance [58]. For example, we consider dataset size to assess whether the augmentation method performs better on smaller or larger datasets. Balancing class distribution is also crucial in evaluating data since unbalanced datasets can degrade model performance.

All datasets chosen for our experiments include modern standard Arabic and multi-dialect data, increasing the model's flexibility and generality when dealing with new data. However, since the Arabic language is morphologically rich, with one word having multiple meanings and shapes, providing diverse Arabic dialects and random examples can affect the proposed approach's performance [9], [27].

Therefore, we selected multiple datasets to cover various aspects of the Arabic language [49], [51], [57]. We chose different sentiment Arabic datasets to experiment and evaluate the proposed approach [59], as listed in Table 3.

However, it is important to acknowledge that the selected datasets, although valuable for our study, represent only a subset of the vast diversity of the Arabic language. Therefore, further research is needed, involving additional datasets with larger sizes and experiments on a wider range of data to strengthen the generalizability of the proposed methodology and provide more comprehensive insights into its performance.

### B. MAIN EXPERIMENT PARAMETERS

Since this study focuses on augmenting Arabic text, we fine-tuned base AraGPT-2 parameters used in [32] for the text generation task and AraBERT parameters used in [33] for the text classification task. Furthermore, two significant decisions were made regarding the datasets' augmentation: (i) selecting the imbalanced class label for any given dataset in the selected datasets to be augmented [15] and (ii) setting the augmentation similarity threshold to the average similarity calculated between the original Arabic text and the generated text of all records in the given dataset from the selected datasets. Table 4 summarizes the augmented classes for each dataset and the average similarity measures considered in this paper.

### C. EXPERIMENTAL ENVIRONMENT AND HARDWARE

The experiment development, implementation, running, and analysis were conducted on an ASUS ROG G703GX notebook. Such a machine runs Windows 10 and has an 8th generation core i9 processor, 64 GB of memory, 2 × 1 TB NVMe SSD RAID hard disk, and NVIDIA GeForce RTX 2080 8 GB graphic card. Given that running the medium and large AraBERT transformer models [33] requires more resources, we ran the base transformer types. To ensure clarity and reproducibility, our code implementation, developed using Python 3.8.10, can be found at [60].

### D. DATASET AUGMENTATION AND GROWTH PERCENTAGE

Our goal is to leverage the transformers' ability to paraphrase Arabic text, and this experiment aims to evaluate the proposed approach's correctness and validity in data modeling and processing techniques. To achieve this, each selected dataset is first preprocessed using the AraBERT preprocessor [33] and then fed to the transformer to generate the corresponding Arabic text. The word embedding is then calculated for the original and generated text to prepare for the similarity calculation [54]. Using the computed average for the selected similarity measure (i.e., Euclidean, cosine, Jaccard, or BLEU similarity) and considering the class labels with fewer instances in the dataset, we start the process of augmenting the given dataset. The final growth percentage is calculated based on the total number of original instances in that set. Tables 5, 6, 7, and 8 summarize the results of this experiment for each dataset in terms of growth counts.

### E. CLASSIFICATION PERFORMANCE AND SIMILARITY PREFERENCE

As mentioned earlier, the proposed approach relies on the similarity threshold, which is calculated by averaging the similarities of all records in the same dataset type. To assess the effectiveness of this approach and validate our assumptions, we conducted five different sentiment classification tasks on both the augmented and original datasets. The first task involved running the selected classifier on the original dataset. For the second, third, fourth, and fifth tasks, we used the AraBERT classifier [33] on datasets resulting from augmentation using the Euclidean, cosine, Jaccard, and BLEU similarity measures [17]–[20], respectively. We then compared the classification results with those obtained using the original dataset, enabling us to evaluate the impact of the augmentation process. The results of these experiments are summarized in Tables 9, 10, 11, and 12, and visualized in Figures 7 and 8.

Additionally, we employed well-established evaluation metrics to comprehensively evaluate the classification performance, namely Receiver Operating Characteristic (ROC) and Precision-Recall (PR) curves. The ROC curves illustrate the trade-off between true positive and false positive rates, while the PR curves demonstrate the relationship between precision and recall. These curves, presented in Figures 9, 10, 11, and 12, showcase the classification performance for the augmented and non-augmented datasets across the AraSarcasem, ASTD, ATT, and MOVIE datasets.

Furthermore, to provide evidence supporting the classification performance results presented in Tables 9-12, we conducted a statistical test, known as the paired t-test [61]. This test is used to determine the significance of the F1 scores for the datasets before and after augmentation. The purpose is to ascertain the statistical significance of the conclusions drawn from these results. We selected a confidence level of 0.05 for this analysis. The respective results for the datasets can be found in Tables 13-16. Tables 13-16 clearly show that all results are statistically significant. These results provide valuable insights into the performance and effectiveness of the proposed approach in sentiment classification tasks across the evaluated datasets.

### F. RESULTS DISCUSSION

This section started by conducting two experiments to validate the proposed methodology. The first experiment was designed to (i) evaluate the validity of using transformer-based models in processing and generating Arabic text and (ii) evaluate the percent of growth for each augmented similarity-based (Euclidean, cosine, Jaccard, and BLEU) dataset on the different selected experimental datasets (AraSarsacm, ASTD, ATT, MOVIE), which have different sizes, labels, and number of instances per labels. The second experiment was conducted to assess the ability of the proposed augmentation approach in enhancing the Arabic sentiment classification performance.

Our results confirm that using Arabic transformer-based models can greatly enhance learning performance in process-

**TABLE 5.** Growth counts (Ara-Sarcasm dataset).

| Class Labels | Dataset Type | | | | |
|---|---|---|---|---|---|
| | Original | Euclidean | cosine | Jaccard | BLEU |
| NEGATIVE | 3528 | 5245 | 4846 | 5317 | 5275 |
| NEUTRAL | 5339 | 5339 | 5339 | 5339 | 5339 |
| POSITIVE | 1678 | 2607 | 2262 | 2459 | 2672 |
| Total | 10545 | 13191 | 12465 | 13115 | 13286 |

**TABLE 6.** Growth counts (ASTD dataset).

| Class Labels | Dataset Type | | | | |
|---|---|---|---|---|---|
| | Original | Euclidean | cosine | Jaccard | BLEU |
| NEGATIVE | 1640 | 1640 | 1640 | 1640 | 1640 |
| NEUTRAL | 805 | 1074 | 1209 | 1134 | 1238 |
| POSITIVE | 776 | 1072 | 1134 | 1110 | 1237 |
| Total | 3221 | 3786 | 3983 | 3884 | 4151 |

**TABLE 7.** Growth counts (ATT dataset).

| Class Labels | Dataset Type | | | | |
|---|---|---|---|---|---|
| | Original | Euclidean | cosine | Jaccard | BLEU |
| POSITIVE | 81 | 86 | 116 | 128 | 126 |
| NEGATIVE | 2070 | 2070 | 2070 | 2070 | 2070 |
| Total | 2151 | 2156 | 2186 | 2198 | 2196 |

**TABLE 8.** Growth counts (MOVIE dataset).

| Class Labels | Dataset Type | | | | |
|---|---|---|---|---|---|
| | Original | Euclidean | cosine | Jaccard | BLEU |
| POSITIVE | 381 | 381 | 556 | 762 | 617 |
| NEGATIVE | 170 | 170 | 247 | 340 | 291 |
| NEUTRAL | 966 | 966 | 966 | 966 | 966 |
| Total | 1517 | 1517 | 1769 | 2068 | 1874 |

**TABLE 9.** AraSarcasm classification performance.

| Augmentation Type | Testing on Augmented Split | | | | Testing on Not-Augmented Split | | | |
|---|---|---|---|---|---|---|---|---|
| | F1 | Accuracy | Precision | Recall | F1 | Accuracy | Precision | Recall |
| BLEU (all-text) | 0.80 | 0.84 | 0.79 | 0.80 | 0.84 | 0.80 | 0.79 | 0.80 |
| BLEU (new-text) | 0.80 | 0.84 | 0.80 | 0.80 | 0.84 | 0.80 | 0.80 | 0.80 |
| cosine (all-text) | 0.78 | 0.83 | 0.78 | 0.79 | 0.83 | 0.78 | 0.78 | 0.79 |
| cosine (new-text) | 0.79 | 0.83 | 0.79 | 0.79 | 0.83 | 0.79 | 0.79 | 0.79 |
| Euclidean (all-text) | 0.76 | 0.80 | 0.78 | 0.78 | 0.80 | 0.76 | 0.77 | 0.77 |
| Euclidean (new-text) | 0.77 | 0.81 | 0.78 | 0.77 | 0.81 | 0.77 | 0.78 | 0.77 |
| Jaccard (all-text) | 0.77 | 0.81 | 0.77 | 0.77 | 0.81 | 0.77 | 0.77 | 0.77 |
| Jaccard (new-text) | 0.78 | 0.83 | 0.78 | 0.78 | 0.83 | 0.78 | 0.78 | 0.78 |
| original (text) | 0.73 | 0.77 | 0.75 | 0.76 | 0.77 | 0.76 | 0.75 | 0.76 |

**TABLE 10.** ASTD classification performance.

| Augmentation Type | Testing on Augmented Split | | | | Testing on Not-Augmented Split | | | |
|---|---|---|---|---|---|---|---|---|
| | F1 | Accuracy | Precision | Recall | F1 | Accuracy | Precision | Recall |
| BLEU (all-text) | 0.76 | 0.77 | 0.76 | 0.76 | 0.76 | 0.77 | 0.76 | 0.76 |
| BLEU (new-text) | 0.70 | 0.73 | 0.71 | 0.71 | 0.70 | 0.73 | 0.71 | 0.71 |
| cosine (all-text) | 0.75 | 0.76 | 0.76 | 0.75 | 0.75 | 0.76 | 0.76 | 0.75 |
| cosine (new-text) | 0.70 | 0.72 | 0.71 | 0.71 | 0.70 | 0.72 | 0.71 | 0.71 |
| Euclidean (all-text) | 0.76 | 0.76 | 0.76 | 0.76 | 0.76 | 0.76 | 0.76 | 0.76 |
| Euclidean (new-text) | 0.69 | 0.71 | 0.69 | 0.70 | 0.69 | 0.71 | 0.69 | 0.70 |
| Jaccard (all-text) | 0.74 | 0.76 | 0.76 | 0.75 | 0.74 | 0.76 | 0.76 | 0.75 |
| Jaccard (new-text) | 0.68 | 0.70 | 0.69 | 0.68 | 0.68 | 0.70 | 0.69 | 0.68 |
| original (text) | 0.70 | 0.74 | 0.72 | 0.69 | 0.70 | 0.74 | 0.72 | 0.69 |

**TABLE 11.** ATT classification performance.

| Augmentation Type | Testing on Augmented Split | | | | Testing on Not-Augmented Split | | | |
|---|---|---|---|---|---|---|---|---|
| | F1 | Accuracy | Precision | Recall | F1 | Accuracy | Precision | Recall |
| BLEU (all-text) | 0.93 | 0.99 | 0.96 | 0.90 | 0.93 | 0.99 | 0.96 | 0.90 |
| BLEU (new-text) | 0.91 | 0.98 | 0.99 | 0.87 | 0.91 | 0.98 | 0.99 | 0.87 |
| cosine (all-text) | 0.85 | 0.98 | 0.99 | 0.85 | 0.85 | 0.98 | 0.99 | 0.85 |
| cosine (new-text) | 0.85 | 0.98 | 0.99 | 0.85 | 0.85 | 0.98 | 0.99 | 0.85 |
| Euclidean (all-text) | 0.89 | 0.98 | 0.99 | 0.82 | 0.89 | 0.98 | 0.99 | 0.82 |
| Euclidean (new-text) | 0.88 | 0.98 | 0.99 | 0.85 | 0.88 | 0.98 | 0.99 | 0.85 |
| Jaccard (all-text) | 0.93 | 0.99 | 0.97 | 0.90 | 0.93 | 0.99 | 0.97 | 0.90 |
| Jaccard (new-text) | 0.95 | 0.99 | 0.97 | 0.93 | 0.95 | 0.99 | 0.97 | 0.93 |
| original (text) | 0.84 | 0.98 | 0.89 | 0.80 | 0.84 | 0.98 | 0.89 | 0.80 |

**TABLE 12.** MOVIE classification performance.

| Augmentation Type | Testing on Augmented Split | | | | Testing on Not-Augmented Split | | | |
|---|---|---|---|---|---|---|---|---|
| | F1 | Accuracy | Precision | Recall | F1 | Accuracy | Precision | Recall |
| BLEU (all-text) | 0.56 | 0.72 | 0.73 | 0.58 | 0.56 | 0.72 | 0.73 | 0.59 |
| BLEU (new-text) | 0.54 | 0.75 | 0.49 | 0.59 | 0.54 | 0.75 | 0.49 | 0.59 |
| cosine (all-text) | 0.47 | 0.74 | 0.52 | 0.50 | 0.47 | 0.74 | 0.52 | 0.50 |
| cosine (new-text) | 0.47 | 0.74 | 0.52 | 0.50 | 0.47 | 0.74 | 0.52 | 0.50 |
| Euclidean (all-text) | 0.53 | 0.73 | 0.80 | 0.57 | 0.53 | 0.73 | 0.80 | 0.57 |
| Euclidean (new-text) | 0.53 | 0.76 | 0.50 | 0.58 | 0.53 | 0.76 | 0.50 | 0.58 |
| Jaccard (all-text) | 0.60 | 0.76 | 0.74 | 0.63 | 0.60 | 0.76 | 0.74 | 0.63 |
| Jaccard (new-text) | 0.55 | 0.76 | 0.51 | 0.61 | 0.55 | 0.76 | 0.51 | 0.61 |
| original (text) | 0.47 | 0.74 | 0.52 | 0.50 | 0.47 | 0.74 | 0.52 | 0.50 |

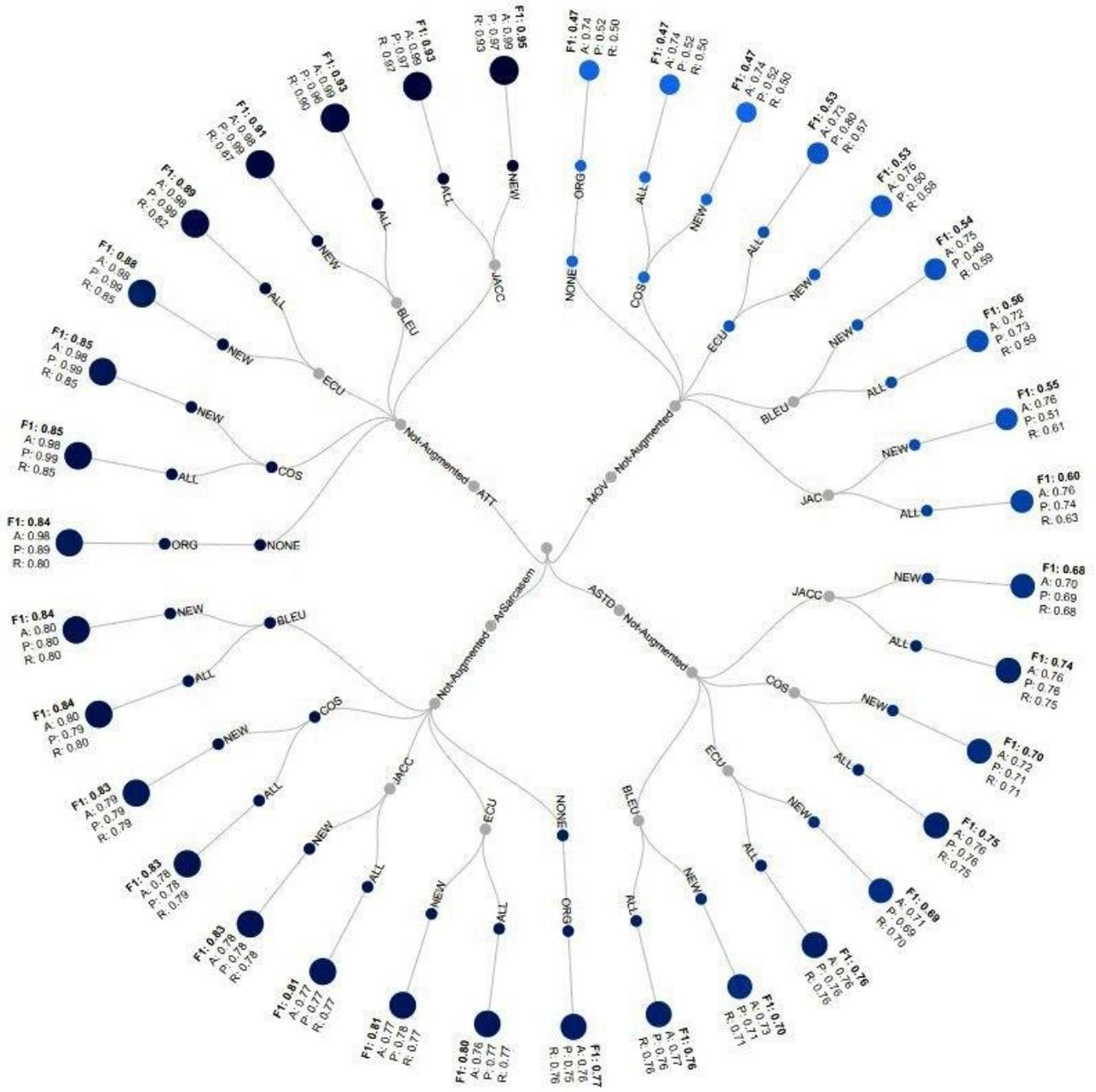

**FIGURE 7.** Sentiment analysis and classification performance results on all data sets (tested on not-augmented split).

**TABLE 13.** Paired t-test results for Arasarcasem Dataset.

| Augmentation Type | Paired t-test | P-value | Conclusion |
|---|---|---|---|
| BLEU (all-text) | 3.5 | 0.02 | Significant |
| BLEU (new-text) | 3.16 | 0.03 | Significant |
| cosine (all-text) | 4 | 0.01 | Significant |
| cosine (new-text) | 2.75 | 0.05 | Significant |
| Euclidean (all-text) | 4.49 | 0.01 | Significant |
| Euclidean (new-text) | 2.7 | 0.05 | Significant |
| Jaccard (all-text) | 4.7 | 0.009 | Significant |
| Jaccard (new-text) | 2.9 | 0.04 | Significant |
| original (text) | 2.64 | 0.05 | Significant |

**TABLE 14.** Paired t-test results for ASTD dataset.

| Augmentation Type | Paired t-test | P-value | Conclusion |
|---|---|---|---|
| BLEU (all-text) | 5.7 | 0.004 | Significant |
| BLEU (new-text) | 2.83 | 0.04 | Significant |
| cosine (all-text) | 2.74 | 0.05 | Significant |
| cosine (new-text) | 2.8 | 0.05 | Significant |
| Euclidean (all-text) | 3.16 | 0.03 | Significant |
| Euclidean (new-text) | 4 | 0.02 | Significant |
| Jaccard (all-text) | 3.5 | 0.02 | Significant |
| Jaccard (new-text) | 3.2 | 0.03 | Significant |
| original (text) | 3.25 | 0.03 | Significant |

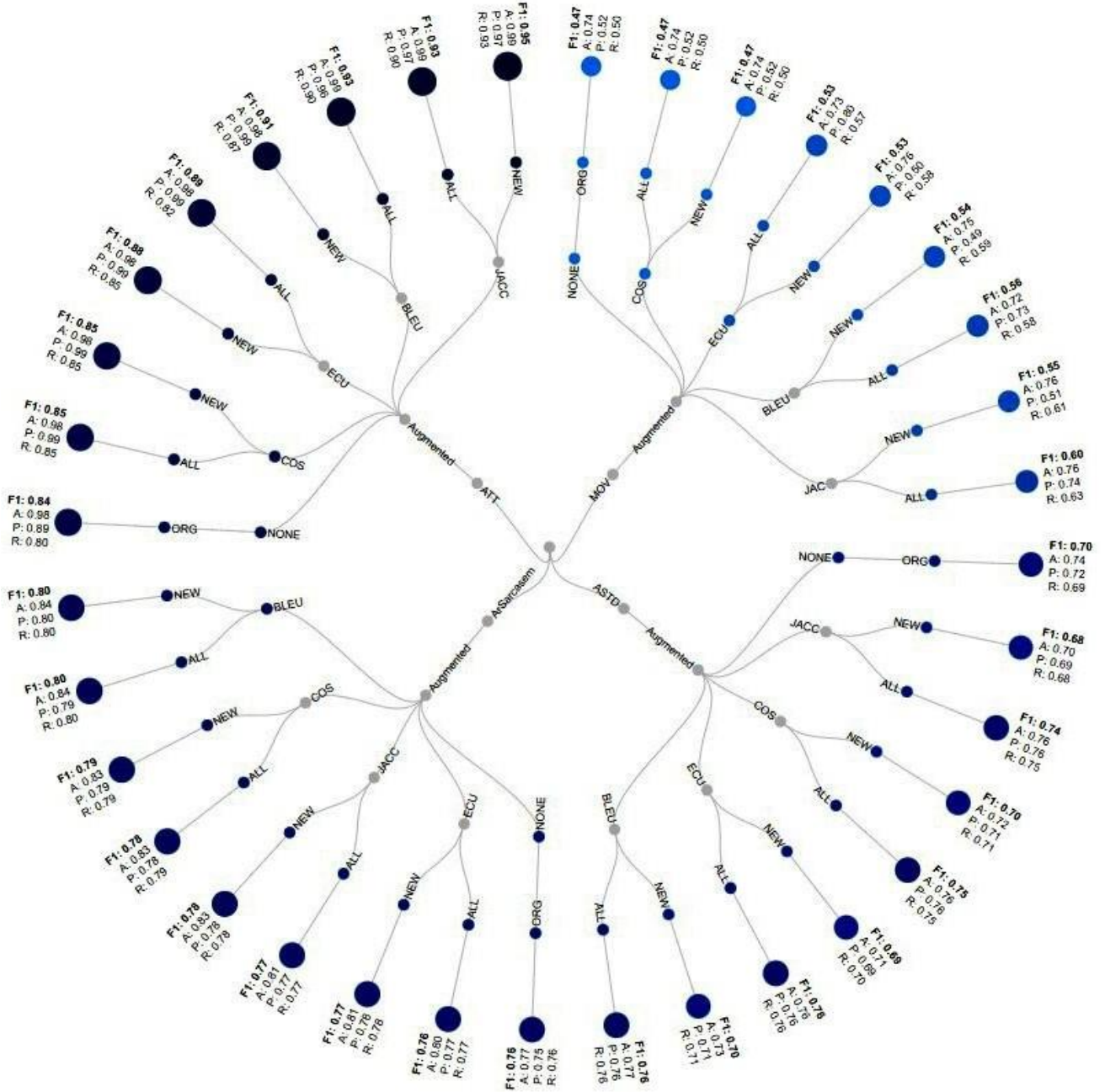

**FIGURE 8.** Sentiment analysis and classification performance results on all data sets (tested on augmented split).

**TABLE 15.** Paired t-test results for ATT dataset.

| Augmentation Type | Paired t-test | P-value | Conclusion |
|---|---|---|---|
| BLEU (all-text) | 3.21 | 0.03 | Significant |
| BLEU (new-text) | 3.5 | 0.02 | Significant |
| cosine (all-text) | 2.95 | 0.04 | Significant |
| cosine (new-text) | 2.76 | 0.05 | Significant |
| Euclidean (all-text) | 3.2 | 0.03 | Significant |
| Euclidean (new-text) | 4 | 0.01 | Significant |
| Jaccard (all-text) | 4.82 | 0.01 | Significant |
| Jaccard (new-text) | 3.77 | 0.02 | Significant |
| original (text) | 3.19 | 0.03 | Significant |

**TABLE 16.** Paired t-test results for MOV dataset.

| Augmentation Type | Paired t-test | P-value | Conclusion |
|---|---|---|---|
| BLEU (all-text) | 5.72 | 0.004 | Significant |
| BLEU (new-text) | 6 | 0.003 | Significant |
| cosine (all-text) | 3.21 | 0.03 | Significant |
| cosine (new-text) | 3.08 | 0.04 | Significant |
| Euclidean (all-text) | 2.83 | 0.04 | Significant |
| Euclidean (new-text) | 2.75 | 0.05 | Significant |
| Jaccard (all-text) | 4.81 | 0.008 | Significant |
| Jaccard (new-text) | 3.2 | 0.03 | Significant |
| original (text) | 4.47 | 0.01 | Significant |

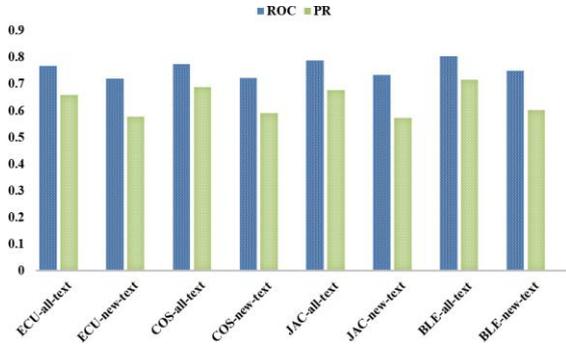

(a) ROC and PR curves for the augmented dataset.

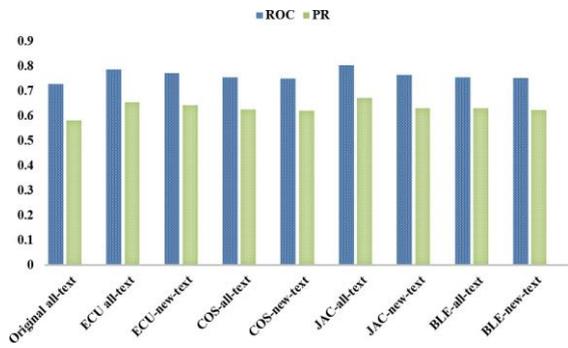

(b) ROC and PR curves for the non-augmented dataset.

**FIGURE 9.** ROC and PR curves for Ara-Sarcasem dataset.

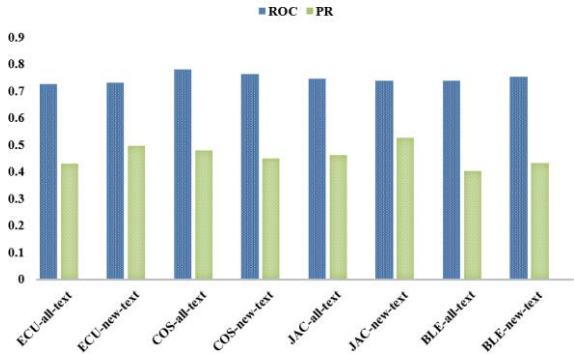

(a) ROC and PR curves for the augmented dataset.

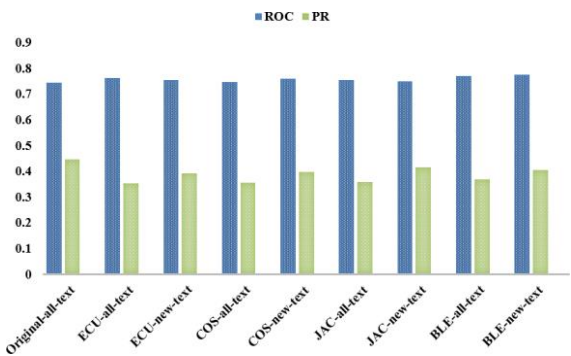

(b) ROC and PR curves for the non-augmented dataset.

**FIGURE 10.** ROC and PR curves for ASTD dataset.

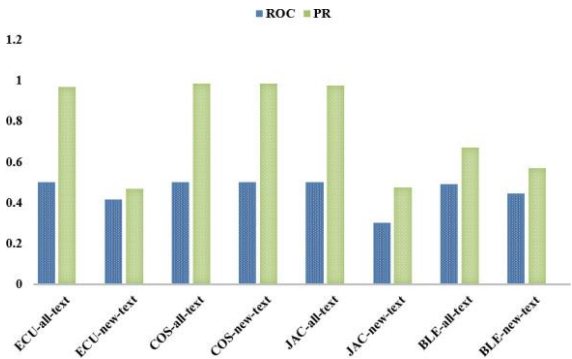

(a) ROC and PR curves for the augmented dataset.

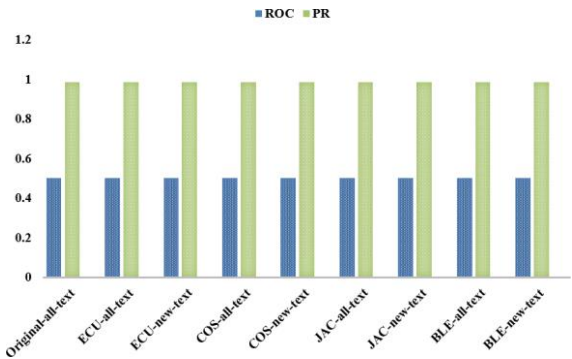

(b) ROC and PR curves for the non-augmented dataset.

**FIGURE 11.** ROC and PR curves for ATT dataset.

ing and generating Arabic text, as expected. However, the results of text generated using AraGPT-2 for augmentation varied across different cases. While some cases yielded perfect text related to each other, other cases produced poor text. This variability can be attributed to the fact that transformer models heavily rely on the accuracy of the data used for pretraining, which is not always correct and accurate for Arabic language models [33]. Despite this limitation, Arabic transformer-based models generally perform well in preserving the context of generated text. Further discussions on the limitations and potential improvements for these models are warranted to gain a deeper understanding of their performance [10].

Further, our results indicate that each similarity-based augmented dataset (using Euclidean, cosine, Jaccard, and BLEU metrics) exhibits a different percentage of growth, which tends to increase with larger datasets and decrease with smaller datasets. Specifically, the AraSarsacm dataset [49] experienced significant growth with +2714 new instances, while the ATT and MOVIE datasets [51] only achieved minor growth with +47 and +551 new instances, respectively, as shown in Tables 5, 6, and 8. This variance in growth can

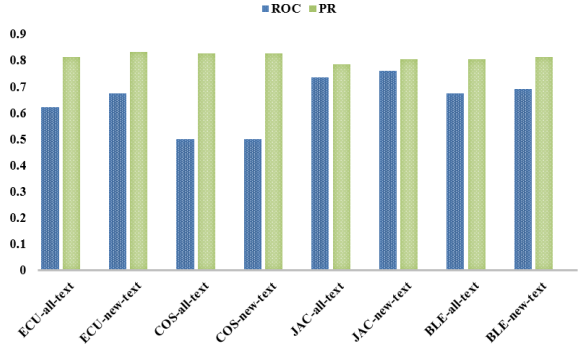

(a) ROC and PR curves for the augmented dataset.

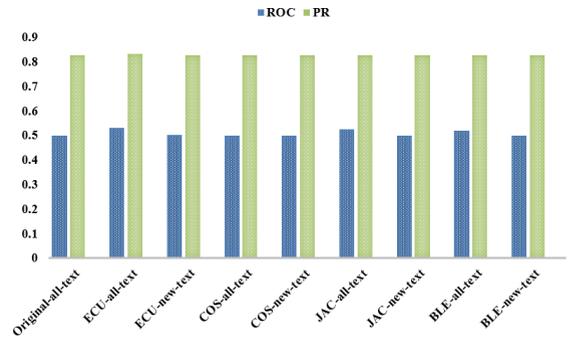

(b) ROC and PR curves for the non-augmented dataset.

**FIGURE 12.** ROC and PR curves for MOVIE dataset.

be attributed to the dataset size itself, as working with larger datasets increases the likelihood of generating new instances, whereas working with smaller datasets limits this potential.

It is noteworthy that the BLEU augmented dataset tends to exhibit high growth percentages in all large datasets, while the Jaccard augmented dataset shows high growth percentages in all small datasets. In contrast, the cosine augmented dataset generally exhibits lower growth percentages across all datasets, except for AraSarsacm, whereas the Euclidean augmented dataset shows different growth percentages across all datasets. The cause of these variations in growth percentages among different similarity metrics can be attributed to different factors, including the sensitivity of the metric to the magnitude of the original and generated sentences, the dataset size, and the nature of the data. All of these factors influence the percentage of growth in each similarity-based augmented dataset. Another finding of the first experiment is that the similarity thresholds for BLEU, Jaccard, and Euclidean were generally lower than those for cosine. The cosine similarity metric tended to score higher threshold percentages, likely due to its calculation being unaffected by sentence size, in contrast to the other similarity metrics, which tend to be influenced by sentence size.

Finally, we note that there appears to be a relationship between the percentage of growth and the enhancement of classification performance. As the growth percentage increases, the classification performance tends to improve. This finding is consistent with previous research on sarcasm and sentiment analysis. Overall, these results highlight the complex interplay between dataset size, similarity metrics, and the percentage of growth in augmented datasets. Further research is needed to explore these relationships in greater depth and identify strategies to optimize the performance of augmented datasets in NLP tasks.

Our second experiment provided further confirmation that the proposed methodology, based on transformers and augmented similarity-based datasets, can effectively enhance Arabic sentiment classification performance. To validate this claim, we compared our proposed augmentation methodology with related studies in the literature [30] using the same dataset [49]. The results of this comparison demonstrated the effectiveness of our approach.

Our findings also support the hypothesis that the performance of learning models improves as the size of the data increases. Specifically, we observed a relationship between the percentage of growth in augmented datasets and the corresponding improvement in classification performance. As a result, the learning model performs better with higher growth percentages in augmented similarity-based datasets.

In addition to these overall findings, we also observed several unexpected results related to the percentage of growth and the use of augmented similarity-based datasets with all text and new text. Some of these observations provide insights into the best similarity metric to use when augmenting imbalanced Arabic datasets. Specifically, we found that the BLEU similarity metric achieved the highest classification performance in all large datasets, while the Jaccard similarity metric was preferred for small datasets, as it achieved the highest classification performance in this context.

Overall, our research provides important insights into the use of augmented similarity-based datasets to enhance Arabic sentiment classification performance. We believe that these findings have important implications for the development of more effective NLP strategies.

## VI. CONCLUSIONS AND FUTURE RESEARCH

Motivated by the power of textual data augmentation (DA) in enhancing text classification, in this paper, we proposed a new DA technique for Arabic text classification, incorporating the unique characteristics of the Arabic language.

In contrast to the existing Arabic DA techniques, which rely only on traditional augmentation methods, our technique employs Arabic transformers to improve DA. Specifically, the AraGPT-2 and AraBERT transformers are exploited in our technique for Arabic text generation and preprocessing, respectively. Furthermore, our technique is designed to utilize several well-known similarity measures, such as the Euclidean, cosine, Jaccard, and BLEU measures, to assess the quality of augmented sentences from different aspects, including context, semantics, and diversity.

We conducted several experiments to assess the effectiveness of our technique in improving Arabic text classification. Our results clearly demonstrated (i) the gains of employing transformer-based models in processing and augmenting imbalanced Arabic datasets and (ii) the powerful impact of combining the cosine, Euclidean, Jaccard, and BLEU similarity measures in preserving the semantics, novelty, and diversity of the augmented sentences. The gains provided by our proposed technique vary depending on the dataset size and the similarity measures growth percent. Our results confirm that BLEU is the preferred similarity metric to augment large imbalanced Arabic datasets, whereas Jaccard is the preferred metric to use when working with small datasets. Our experiments, conducted on different datasets with distinct characteristics such as dataset size, label number, and unbalanced classes, showed significant improvement in sentiment classification performance compared to existing techniques [30].

Finally, addressing the limitations and potential shortcomings of the proposed method is crucial for achieving a balanced perspective on its effectiveness and applicability. In the following, we discuss the identified limitations of our study. Firstly, the variability in the performance of the proposed approach when tested on different datasets is acknowledged, requiring careful consideration and further investigation. Secondly, the utilization of basic transformer models, due to limited hardware resources, may have constrained the potential gains of the technique. It is essential to highlight that the approach's performance can be further improved if more advanced transformer models can be utilized, as suggested in [9]. Moreover, in future research, analyzing various data types and their impact on the performance of our proposed method is important. Additionally, conducting a thorough investigation and evaluation of diverse datasets is necessary to gain a deeper understanding of the limitations and opportunities for improvement within the proposed approach. By addressing these limitations and exploring these avenues, a more comprehensive and balanced perspective on the effectiveness and applicability of the proposed approach can be reached.